# Argument Calculus and Networks

Adnan Y. Darwiche
Cognitive Systems Laboratory
Computer Science Department
University of California
Los Angeles, CA 90024
*darwiche@cs.ucla.edu*

## Abstract

A major reason behind the success of probability calculus is that it possesses a number of valuable tools, which are based on the notion of probabilistic independence. In this paper, I identify a notion of logical independence that makes some of these tools available to a class of propositional databases, called argument databases. Specifically, I suggest a graphical representation of argument databases, called argument networks, which resemble Bayesian networks. I also suggest an algorithm for reasoning with argument networks, which resembles a basic algorithm for reasoning with Bayesian networks. Finally, I show that argument networks have several applications: Nonmonotonic reasoning, truth maintenance, and diagnosis.

## 1   INTRODUCTION

A major reason behind the success of probability calculus is that it possesses a number of valuable tools, which are based on the notion of probabilistic independence [Pearl, 1988]. In this paper, I identify an intuitive notion of logical independence that makes some of these tools available to a special class of propositional databases.

In particular, I identify in Section 2 a class of propositional databases, called argument databases, and study some of their properties. In Section 3, I identify a notion of logical independence with respect to argument databases and study its properties. In Section 4, I suggest a graphical representation of argument databases, called argument networks, which resemble Bayesian networks. And in Section 5, I suggest an algorithm for reasoning with argument networks, which resembles a basic algorithm for reasoning with Bayesian networks. Finally, I show in Section 6 that argument networks have several applications: Nonmonotonic reasoning, truth maintenance, and diagnosis. Proofs, omitted due to space limitations, can be found in the full version of this paper.

## 2   ARGUMENT DATABASES

Logical independence, to be defined in Section 3, is based on three notions: argument databases, arguments, and conditional arguments, which are counterparts of probability distributions, probabilities, and conditional probabilities. This section explores these three notions in some detail.

**Definition 1** *Let $\mathcal{L}$ and $\mathcal{A}$ be two propositional languages over disjoint primitive propositions. An argument database $\Delta$ with respect to $(\mathcal{L}, \mathcal{A})$ is a set of sentences $\alpha \supset \phi$, where sentence $\alpha$ belongs to language $\mathcal{A}$, sentence $\phi$ belongs to language $\mathcal{L}$, and database $\Delta$ does not entail any invalid sentence in language $\mathcal{A}$.*[1]

**Example 1** Let $\mathcal{L}$ be a propositional language constructed from primitive propositions *rain*, *sprinkler_on*, *wet_grass*, and *wet_shoes*. Let $\mathcal{A}$ be another propositional language constructed from primitive propositions $a_1, \ldots, a_6$. The following is an argument database with respect to $(\mathcal{L}, \mathcal{A})$:

$$\Delta = \begin{array}{ll} a_1 & \supset \quad rain \\ a_2 & \supset \quad sprinkler\_on \\ a_3 & \supset \quad (rain \supset wet\_grass) \\ a_4 & \supset \quad (sprinkler\_on \supset wet\_grass) \\ a_5 & \supset \quad wet\_grass \\ a_6 & \supset \quad (wet\_grass \supset wet\_shoes). \end{array}$$

### 2.1   Arguments

The same way that a probability distribution assigns a unique probability to each sentence, an argument database assigns a unique argument (up to logical equivalence) to every sentence:

---

[1] Any propositional database is an argument database with respect to some pair $(\mathcal{L}, \mathcal{A})$.

**Definition 2** *Let $\Delta$ be an argument database with respect to $(\mathcal{L}, \mathcal{A})$ and let $\phi$ be a sentence in $\mathcal{L}$. The argument for sentence $\phi$ with respect to database $\Delta$, written $\Delta(\phi)$, is the weakest sentence $\alpha$ in language $\mathcal{A}$ that together with database $\Delta$ entails sentence $\phi$: $\Delta \cup \{\alpha\} \models \phi$.*[2]

Any sentence in $\mathcal{A}$ that entails $\Delta(\phi)$ is called <u>an</u> argument for $\phi$. Recall that $\Delta(\phi)$ itself is <u>the</u> argument for $\phi$.

As we shall see later, the argument for a sentence is closely related to the ATMS label of the sentence [Reiter and de Kleer, 1987]. In particular, I will show in Section 6 that the prime implicants for the argument $\Delta(\phi)$ constitute the label for the sentence $\phi$.

**Example 2** Consider Example 1. The argument for *wet_grass*, $\Delta(wet\_grass)$, is $(a_1 \wedge a_3) \vee (a_2 \wedge a_4) \vee a_5$. Moreover, each of $a_1 \wedge a_3$, $a_2 \wedge a_4$, and $a_5$ is an argument for *wet_grass*.

Properties of argument databases are similar to properties of probability distributions:

**Theorem 1** *An argument database $\Delta$ satisfies:*

1. $\Delta(\text{true}) \equiv \text{true}$,

2. $\Delta(\text{false}) \equiv \text{false}$,

3. $\Delta(\phi \wedge \psi) \equiv \Delta(\phi) \wedge \Delta(\psi)$, *and*

4. $\Delta(\phi) \equiv \Delta(\psi)$ *when* $\phi \equiv \psi$.

Note how **true** and **false** in *argument calculus* play the roles of 1 and 0 in probability calculus.

Although the argument for a conjunction can be computed from the arguments for its conjuncts, the argument for a disjunction cannot be computed from the arguments for its disjuncts in general:

**Theorem 2** $\Delta(\phi) \vee \Delta(\psi) \models \Delta(\phi \vee \psi)$, *but* $\Delta(\phi \vee \psi) \not\models \Delta(\phi) \vee \Delta(\psi)$.

**Example 3** Consider the argument database $\{a_3 \supset (rain \supset wet\_grass)\}$. The argument for $\neg rain$ is **false**, the argument for *wet_grass* is **false**, but the argument for $\neg rain \vee wet\_grass$ is $a_3$.

The role that conjunction and disjunction play in argument calculus is dual to the role they play in probability calculus. In probability calculus, the probability of a disjunction can be computed from the probabilities of the disjuncts when the disjuncts are logically disjoint. However, to compute the probability of a conjunction one has to appeal to the notion of conditional probability unless the conjuncts are independent. In argument calculus, however, the argument for a conjunction can be computed from the arguments for the conjuncts[3], but to compute the argument for a disjunction one has to appeal to the notion of conditional argument unless the conjuncts are independent. Conditional arguments and independence shall be discussed next.

## 2.2 Conditional arguments

The obvious way to update the argument for $\psi$ after observing some sentence $\phi$ in $\mathcal{L}$ is to compute the argument for $\psi$ with respect to the extended database $\Delta \cup \{\phi\}$. This computation gives the argument for $\phi \supset \psi$ with respect to the database $\Delta$. But this argument includes the argument for $\neg\phi$, which should not count because $\phi$ has been observed. When the argument for $\neg\phi$ is subtracted from the argument for $\phi \supset \psi$, we get the conditional argument for $\psi$ given $\phi$.

**Definition 3** *The <u>conditional argument</u> for $\psi$ given $\phi$, written $\Delta(\psi \mid \phi)$, is*

$$\Delta(\psi \mid \phi) \stackrel{def}{=} \Delta(\phi \supset \psi) \wedge \neg\Delta(\neg\phi).$$

**Example 4** Consider the argument database $\{a_1 \supset rain\}$. The argument for $\neg rain \supset wet\_grass$ is $a_1$, which is also the argument for $\neg rain$. The conditional argument for *wet_grass* given $\neg rain$ is $a_1 \wedge \neg a_1 \equiv$ **false**. Therefore, although there is an argument for $\neg rain \supset wet\_grass$, there is no argument for *wet_grass* given $\neg rain$.

Although conditional arguments play a central role in defining logical independence, a related class of arguments, called sufficient arguments, plays a central role in computing arguments.

**Definition 4** *A <u>sufficient argument</u> for $\psi$ given $\phi$, written $\Delta(\phi \rightarrow \psi)$, is an argument that satisfies $\Delta(\psi \mid \phi) \equiv \Delta(\phi \rightarrow \psi) \models \Delta(\psi \supset \phi)$.*

A sufficient argument for $\psi$ given $\phi$ is "sufficient" for computing the argument for $\phi \supset \psi$ once the argument for $\neg\phi$ is computed:

**Theorem 3 (Disjunction Rule)** $\Delta(\phi \supset \psi) \equiv \Delta(\phi \rightarrow \psi) \vee \Delta(\neg\phi)$.

**Example 5** Consider the argument database $\{a_7 \supset \neg rain, a_3 \supset (rain \supset wet\_grass)\}$. The argument for $rain \supset wet\_grass$ is $a_3 \vee a_7$ and the argument for *wet_grass* given *rain* is $a_3 \wedge \neg a_7$. It follows that $a_3$ is a sufficient argument for *wet_grass* given *rain*. Therefore, disjoining $a_3$ with the argument for $\neg rain$ gives the argument for $rain \supset wet\_grass$.

---

[2]Sentence $\alpha$ is weaker than sentence $\beta$ if $\beta$ entails $\alpha$. The argument for a sentence is unique up to logical equivalence.

[3]The equivalence $\Delta(\phi \wedge \psi) \equiv \Delta(\phi) \wedge \Delta(\psi)$ holds even when the conjuncts $\phi$ and $\psi$ are not logically disjoint. This is because logical conjunction is idempotent; that is, $\alpha \wedge \alpha \equiv \alpha$ for all $\alpha$, which is not true of numeric addition since $a + a \neq a$ in general.

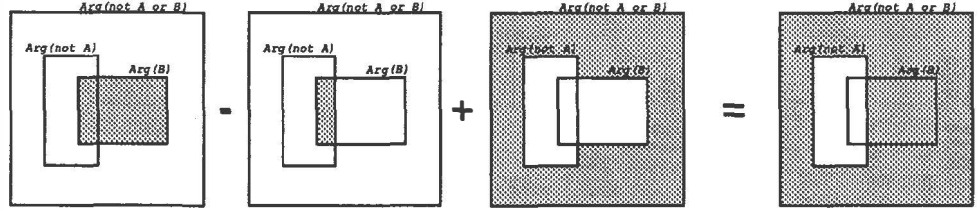

Figure 1: ($A$ stands for $\phi$ and $B$ stands for $\psi$ ). The change that occurs to the argument for $B$ as a result of observing $A$. From left to right, the above shaded areas are: the argument for $B$, the negative influence of $A$ on $B$, the positive influence of $A$ on $B$, and the conditional argument for $B$ given $A$.

## 3   INDEPENDENCE

The notion of logical independence is based on the relation between arguments and conditional arguments. Consider Figure 1, for example, which depicts the relation between the argument for $\psi$ and the conditional argument for $\psi$ given $\phi$. The two arguments are incomparable in general. The decrease in the argument for $\psi$ after observing $\phi$ is called the negative influence of $\phi$ on $\psi$. And the increase in the argument for $\psi$ after observing $\phi$ is called the positive influence of $\phi$ on $\psi$. The positive influence of $\phi$ on $\psi$ is the disjunction of all arguments for $\phi \supset \psi$ that are neither arguments for $\neg\phi$ nor arguments for $\psi$. And the negative influence of $\phi$ on $\psi$ is the disjunction of all arguments for $\psi$ that are also arguments for $\neg\phi$. More formally:

**Definition 5** *The positive influence of $\phi$ on $\psi$, written $\Delta(\phi \rightsquigarrow \psi)$, is $\Delta(\phi \supset \psi) \wedge \neg\Delta(\neg\phi) \wedge \neg\Delta(\psi)$. The negative influence of $\phi$ on $\psi$ is $\Delta(\psi \wedge \neg\phi)$.*

**Example 6** Consider the argument database:

$$
\begin{aligned}
a_7 &\supset \neg rain \\
a_5 &\supset wet\_grass \\
a_3 &\supset (rain \supset wet\_grass)
\end{aligned}
$$

The negative influence of $rain$ on $wet\_grass$ is $a_5 \wedge a_7$ because this will be subtracted from the argument for $wet\_grass$ when $rain$ is observed. The positive influence of $rain$ on $wet\_grass$ is $a_3 \wedge \neg a_5 \wedge \neg a_7$ because this will be added to the argument for $wet\_grass$ when $rain$ is observed.

When $\Delta(\phi \rightsquigarrow \psi) \equiv \textbf{false}$, we say that $\phi$ has no positive influence on $\psi$. And when $\Delta(\psi \wedge \neg\phi) \equiv \textbf{false}$, we say that $\phi$ has no negative influence on $\psi$.

Below are two definitions of independence that are based on positive and negative influence. According to the first definition, a set of propositions $I$ is independent from another set $J$ precisely when no information about propositions $J$ has a positive influence on any information about propositions $I$. According to the second definition, $I$ is independent from $J$ precisely when no information about $J$ has a negative influence on any information about $I$.

Before I state the definitions formally, let me introduce some notation. The symbol $\hat{i}$ denotes a literal, $i$ or $\neg i$, where $i$ is a primitive proposition. The symbol $\hat{I}$ denotes a conjunction of literals $\hat{i}$, where $i$ belongs to $I$. And the symbol $\check{I}$ denotes a disjunction of literals $\hat{i}$, where $i$ belongs to $I$.

**Definition 6** *An argument database $\Delta$ finds propositions $I$ +independent from propositions $J$, written +$Ind_\Delta(I, J)$, precisely when no $\hat{J}$ has a positive influence on any $\check{I}$. And $\Delta$ finds propositions $I$ −independent from $J$, written −$Ind_\Delta(I, J)$, precisely when no $\hat{J}$ has a negative influence on any $\check{I}$.*

**Corollary 1** +$Ind_\Delta(I, J)$ iff $\Delta(\check{I} \mid \hat{J}) \models \Delta(\check{I})$ and −$Ind_\Delta(I, J)$ iff $\Delta(\check{I}) \models \Delta(\check{I} \mid \hat{J})$.

**Example 7** Consider Example 1. *sprinkler_on* is +independent of $rain$, but is −dependent on $rain$. Moreover, *wet_shoes* is +dependent on $rain$.

From here on, I will discuss +independence only.

There is also a notion of conditional +independence in argument calculus. It can be defined in terms of conditional influence, but the following is a simpler definition in terms of conditional arguments.

**Definition 7** *An argument database $\Delta$ finds propositions $I$ +independent from $J$ given $K$, written +$Ind_\Delta(I, K, J)$, precisely when*

$$\Delta(\check{I} \mid \hat{K} \wedge \hat{J}) \models \Delta(\check{I} \mid \hat{K}).$$

**Example 8** In Example 1, *wet_shoes* is +independent of $rain$ given $wet\_grass$.

There are several characterizations of conditional +independence in terms of arguments, conditional arguments, and sufficient arguments. Following is one of these characterizations.

**Theorem 4** +$Ind_\Delta(I, K, J)$ iff

$$\Delta(\hat{K} \supset \check{I} \vee \check{J}) \equiv \Delta(\hat{K} \supset \check{I}) \vee \Delta(\hat{K} \supset \check{J}).$$

Of most importance among the properties of conditional +independence are the graphoid axioms [Pearl, 1988]:

**Theorem 5** *Conditional +independence satisfies the following properties:*
*(a) $+Ind_\Delta(I, K, J)$ iff $+Ind_\Delta(J, K, I)$, and*
*(b) $+Ind_\Delta(I, K, J)$ and $+Ind_\Delta(L, K \cup I, J)$ iff $+Ind_\Delta(I \cup L, K, J)$.*

# 4    ARGUMENT NETWORKS

An argument network is a graphical representation of an argument database. Figure 2 depicts an argument network, which represents the database of Example 1. Figure 3 depicts another argument network.

An argument network has two components: a directed acyclic graph and a set of tables. Every node in an argument network has a table associated with it. The table has two columns, each corresponding to a state of the associated node. The table also has a number of rows, each corresponding to a state of the node's parents. A table entry at row $\phi$ and column $\psi$ is an argument for $\phi \supset \psi$. For example, the top left entry of the table associated with Node *wet_grass*, $a_3 \vee a_4 \vee a_5$, is an argument for $rain \wedge sprinkler\_on \supset wet\_grass$.

Following is the formal definition of an argument network in which the symbol $i\diamond$ denotes the parents of node $i$.

**Definition 8** *An argument network is a tuple $\langle \mathcal{L}, \mathcal{A}, \mathcal{G}, \mathcal{Q} \rangle$, where*

1. *$\mathcal{L}$ and $\mathcal{A}$ are propositional languages over disjoint primitive propositions,*

2. *$\mathcal{G}$ is a directed acyclic graph over the primitive propositions of language $\mathcal{L}$, and*

3. *$\mathcal{Q}$ maps each pair $(i\diamond, i)$, where $i$ is a node in $\mathcal{G}$, into an argument in $\mathcal{A}$ such that $\mathcal{Q}(i\diamond, i) \wedge \mathcal{Q}(i\diamond, \neg i) \equiv \mathbf{false}$.*

**Definition 9** *The database corresponding to argument network $\langle \mathcal{L}, \mathcal{A}, \mathcal{G}, \mathcal{Q} \rangle$ is*

$$\{\mathcal{Q}(i\diamond, \hat{i}) \supset (i\diamond \supset \hat{i}) \mid i \text{ is a node in } \mathcal{G}\}.$$

An argument network graphically explicates many of the independences in its corresponding database. The following two theorems elaborate on this and other features.

**Theorem 6** *Let $\langle \mathcal{L}, \mathcal{A}, \mathcal{G}, \mathcal{Q} \rangle$ be an argument network and let $\Delta$ be its corresponding database. Then*

1. *$\Delta$ is an argument database,*

2. *the argument $\mathcal{Q}(i\diamond, \hat{i})$ is a sufficient argument for $\hat{i}$ given $i\diamond$, and*

3. *any node in $\mathcal{G}$ is +independent from its nondescendents given its parents.*

The first result above says that the database corresponding to an argument network does not entail any invalid; sentence in the language $\mathcal{A}$. The second result says that $\mathcal{Q}(i\diamond, \hat{i})$ is entailed by the conditional argument $\Delta(\hat{i} \mid i\diamond)$ and entails the argument $\Delta(i\diamond \supset \hat{i})$. The third result is most interesting because it shows that some independences, which are *part of the definition* of a Bayesian network, are *properties* of an argument network. Together with Theorem 5, this result leads to the following consequential theorem.

**Theorem 7** *Let $\langle \mathcal{L}, \mathcal{A}, \mathcal{G}, \mathcal{Q} \rangle$ be an argument network and let $I, J, K$ be disjoint sets of nodes in $\mathcal{G}$. If $K$ d–separates $I$ from $J$, then $+Ind_\Delta(I, K, J)$.*

The criterion of $d$–separation is a topological test that is not defined here, but can be found elsewhere [Pearl, 1988].

# 5    COMPUTING ARGUMENTS

A basic algorithm for computing probabilities in Bayesian networks is the well known *polytree algorithm* [Pearl, 1988; Peot and Shachter, 1991]. Although this algorithm applies to singly connected networks,[4] it can be extended to multiply connected networks [Horvitz *et al.*, 1989; Pearl, 1988; Suermondt and Cooper, 1988; Peot and Shachter, 1991]. In this section, I present a similar algorithm for computing arguments in singly connected networks, which can be extended to compute arguments in multiply connected networks [Darwiche, 1992].

Given an observation $\delta$, the algorithm computes the argument $\Delta(\delta \supset i)$ for each literal $\hat{i}$. From such arguments, one computes the argument for the negated observation $\neg \delta$ using $\Delta(\neg \delta) \equiv \Delta(\delta \supset i) \wedge \Delta(\delta \supset \neg i)$. Then one computes the conditional argument for $\hat{i}$ given $\delta$ using $\Delta(\hat{i} \mid \delta) \equiv \Delta(\delta \supset \hat{i}) \wedge \neg \Delta(\neg \delta)$. In the following theorem, which states the algorithm, the symbol $i\diamond j$ denotes the parents of node $i$ except parent $j$, $i\diamond$ denotes the children of node $i$, and $i\diamond j$ denotes the children of node $i$ except child $j$

**Theorem 8** *Let $\langle \mathcal{L}, \mathcal{A}, \mathcal{G}, \mathcal{Q} \rangle$ be an argument network and let $\Delta$ be its corresponding database. Let $\delta$ be a state of some leaf nodes in $\mathcal{G}$, where each node has only one parent. If $i$ is a non–observed node in $\mathcal{G}$, then $\Delta(\delta \supset \hat{i})$ equals $\pi_i(\hat{i}) \vee \lambda_i(\hat{i})$, where*

$$\pi_i(\hat{i}) \quad \stackrel{def}{=} \quad \bigwedge_{i\diamond} \mathcal{Q}(i\diamond, \hat{i}) \vee \bigvee_{i\diamond \models j} \pi_{j.i}(\neg \hat{j}),$$

---

[4] A singly connected network has only one undirected path between any two nodes.

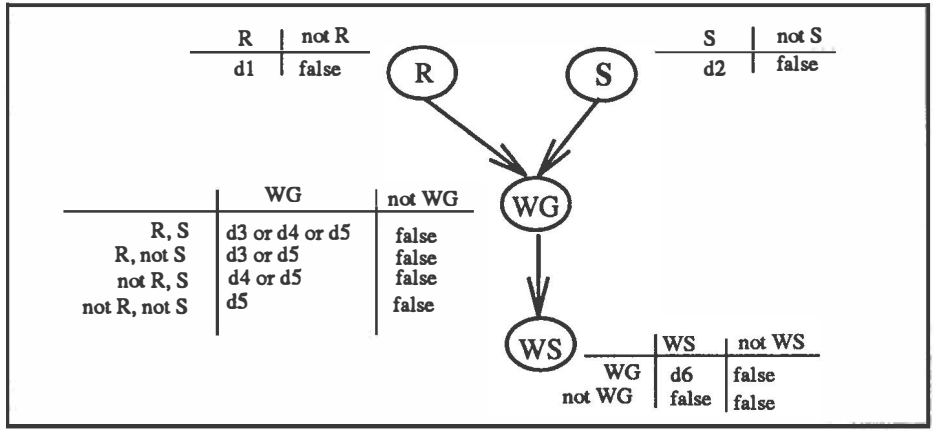

Figure 2: An argument network. The symbols $R, S, WG,$ and $WS$ stand for *rain, sprinkler_on, wet_grass,* and *wet_shoes,* respectively.

$$\lambda_i(\hat{\imath}) \stackrel{def}{=} \bigvee_{k \in io} \lambda_{k.i}(\hat{\imath}),$$

$$\pi_{j.i}(\hat{\jmath}) \stackrel{def}{=} \pi_j(\hat{\jmath}) \vee \bigvee_{k \in joi} \lambda_{k.j}(\hat{\jmath}),$$

$\lambda_{k.i}(\hat{\imath}) \stackrel{def}{=} \mathcal{Q}(\neg \hat{\imath}, \neg k),$ if $\delta \models \hat{k}$; and

$$\bigwedge_{\hat{k}} \lambda_k(\neg \hat{k}) \vee \bigwedge_{k \hat{o}i} \mathcal{Q}(k \hat{o}i \wedge \neg \hat{\imath}, \neg \hat{k}) \vee \bigvee_{k \hat{o}i \models j} \pi_{j.k}(\neg \hat{\jmath}),$$

*otherwise.*

The polytree algorithm is usually explained in terms of a message–passing paradigm in which the pair $\langle \pi_{j.i}(j), \pi_{j.i}(\neg j) \rangle$ is called the message from node $j$ to its child $i$ and the pair $\langle \lambda_{k.i}(i), \lambda_{k.i}(\neg i) \rangle$ is called the message from node $k$ to its parent $i$. The computation of the algorithm is then a sequence of message exchanges between nodes in which each node receives and sends one message to each neighbor. Therefore, the number of messages exchanged during the computation is twice the number of arcs in the network, which, for singly connected networks, is one less than the number of nodes.

Beyond its message–passing behavior, the polytree algorithm is well known for its time complexity. Theorem 9 below shows a similar time complexity for the algorithm of Theorem 8, assuming that constructing a disjunction (or conjunction) of $l$ elements requires $l$ units of space and $l$ units of time.

**Theorem 9** *A non–observed node with $n > 0$ parents and $m > 0$ children consumes $(n+2)2^{n+1} + 2m$ space units and a similar number of time units when it sends a child message, and consumes $(n+1)2^{n+1} + 4(m+2)$ space units and a similar number of time units when it sends a parent message.*

The theorem shows that the time and space consumed by the algorithm is manageable if the number of parents per node is small. In particular, when there is one parent per node (the network is a tree), the time of the algorithm and the size of all arguments constructed are linear in both the number of nodes in the network and the number of children per node.

## 6    APPLICATIONS OF ARGUMENT NETWORKS

In this section, I discuss three applications of argument networks: Nonmonotonic reasoning, truth maintenance, and diagnosis. In nonmonotonic reasoning, I show how to compute what needs to be retracted from a database in order to resolve a conflict with an observation. In truth maintenance, I show how to compute the label of a sentence [Reiter and de Kleer, 1987] from its argument. And in diagnosis, I show how to compute the kernel diagnoses [de Kleer *et al.*, 1992] of an observation from the argument for the negated observation.

All three applications are isomorphic at some level of abstraction. Moreover, in all of them, we end up expressing some argument in its prime implicant form. Following is a review of the notion of a prime implicant and the connected notion of a prime implicate.

**Definition 10** *A* <u>*conjunctive clause*</u> *is a conjunction of literals.* <u>*An implicant*</u> *for sentence $\psi$ is a satisfiable conjunctive clause that entails $\psi$. A* <u>*prime implicant*</u> *for $\psi$ is a weakest implicant for $\psi$. A* <u>*disjunctive clause*</u> *is a disjunction of literals. An* <u>*implicate of sentence*</u> *$\psi$ is an invalid disjunctive clause that is entailed by $\psi$. A* <u>*prime implicate*</u> *of $\psi$ is a strongest implicate of $\psi$.*

### 6.1    Nonmonotonic reasoning

When our beliefs are represented by a propositional database, we are often interested in answering two types of questions. First, does sentence $\phi$ follow from the database? And second, if $\phi$ follows from the database, and if we observe $\neg\phi$, then what should be removed from the database such that the conflict is resolved? Both of these questions can be answered by appealing to the notion of an argument.

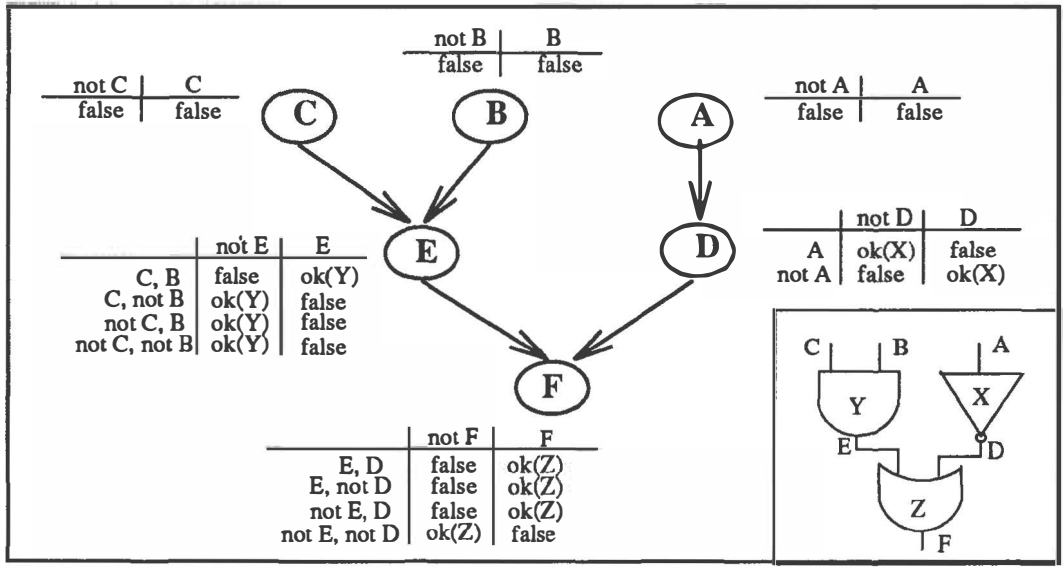

Figure 3: An argument network representing the circuit on the right corner. This network assumes a particular fault model of digital gates: If a gate is OK, it produces the right output; but if the gate is not OK, it may or may not produce the right output.

In particular, suppose that we have a database $\Gamma = \{\phi_1, \ldots, \phi_n\}$ that is constructed from language $\mathcal{L}$. To answer the above questions, we introduce a primitive proposition $a_i$ to represent the identity of each sentence $\phi_i$ in the database — the argument language $\mathcal{A}$ is constructed from these primitive propositions. We then construct the argument database $\Delta = \{a_1 \supset \phi_1, \ldots, a_n \supset \phi_n\}$. For example, the database

$$\Gamma = \begin{array}{l} rain \\ sprinkler\_on \\ rain \supset wet\_grass \\ sprinkler\_on \supset wet\_grass \\ wet\_grass \\ wet\_grass \supset wet\_shoes. \end{array}$$

gets represented by the argument database:

$$\Delta = \begin{array}{lll} a_1 & \supset & rain \\ a_2 & \supset & sprinkler\_on \\ a_3 & \supset & rain \supset wet\_grass \\ a_4 & \supset & sprinkler\_on \supset wet\_grass \\ a_5 & \supset & wet\_grass \\ a_6 & \supset & wet\_grass \supset wet\_shoes. \end{array}$$

The argument network of this database was given in Figure 2.

The database $\Gamma$ entails some sentence $\phi$ precisely when $\Delta \cup \{a_1 \wedge \ldots \wedge a_n\} \models \phi$. And this holds precisely when $a_1 \wedge \ldots \wedge a_n$ entails the argument $\Delta(\phi)$, which can be tested in time proportional to the size of the argument $\Delta(\phi)$.

When the observation $\phi$ is inconsistent with the database $\Gamma$, one is usually interested in retracting a set of sentences from the database $\Gamma$ to make it consistent with the observation $\phi$. There is often more than one set of sentences that can achieve this, and the prime implicants for the negated argument $\neg\Delta(\neg\phi)$ characterize all of them. In particular, the negative literals of a prime implicant for $\neg\Delta(\neg\phi)$ correspond to a minimal set of sentences that must be retracted, and its positive literals correspond to a minimal set of sentences that must not be retracted, in order for the database $\Gamma$ to become consistent with the observation $\phi$.

**Example 9** Consider the database $\Gamma$ above and its corresponding argument database $\Delta$. We want to know whether $\Gamma$ entails $wet\_grass$. We can answer this question by answering another question: Does $\Delta \cup \{a_1, \ldots, a_6\}$ entail $wet\_grass$? To answer this question, we compute the argument for $wet\_grass$ and test whether $a_1 \wedge \ldots \wedge a_6$ entails it. The argument for $wet\_grass$ was computed in Example 2 to be $(a_1 \wedge a_3) \vee (a_2 \wedge a_4) \vee a_5$. This argument is entailed by $a_1 \wedge \ldots \wedge a_6$. Therefore, $\Gamma$ entails $wet\_grass$. Now, suppose that we observe $\neg wet\_grass$, which contradicts the database $\Gamma$. What should be retracted from $\Gamma$ to resolve this contradiction? To answer this question, we compute the prime implicants for $\neg\Delta(\neg wet\_grass)$, which turn

out to be:

$$\neg a_1 \wedge \neg a_2 \wedge \neg a_5$$
$$\neg a_1 \wedge \neg a_4 \wedge \neg a_5$$
$$\neg a_3 \wedge \neg a_2 \wedge \neg a_5$$
$$\neg a_3 \wedge \neg a_4 \wedge \neg a_5.$$

Each one of these implicants characterize a minimal set of sentences that must be retracted from $\Gamma$ in order to resolve the conflict with the given observation. For example, the first implicant says that if we remove *rain*, *sprinkler_on* and *wet_grass* from $\Gamma$, then $\neg wet\_grass$ will no longer be inconsistent with the resulting $\Gamma$:

$$rain \supset wet\_grass$$
$$sprinkler\_on \supset wet\_grass$$
$$wet\_grass \supset wet\_shoes.$$

The fourth implicant, however, says that if we remove *rain* $\supset$ *wet_grass*, *sprinkler_on* $\supset$ *wet_grass*, and *wet_grass* from $\Gamma$, then $\neg wet\_grass$ will no longer be inconsistent with the resulting $\Gamma$:

$$rain$$
$$sprinkler\_on$$
$$wet\_grass \supset wet\_shoes.$$

And so on.

## 6.2    Truth maintenance

The basic task of an assumption–based truth maintenance system, also called a clause management system (CMS) [Reiter and de Kleer, 1987], is to compute labels of sentences. Roughly speaking, the label for a sentence is a set of "minimal" arguments for that sentence. More formally, we have the following definitions [Reiter and de Kleer, 1987]:

**Definition 11** *A minimal support for sentence $\phi$ with respect to database $\Delta$ is a prime implicate of $\Delta \cup \{\neg\phi\}$ that is not an implicate of $\Delta$.*

**Definition 12** *The $\underline{\mathcal{A}\text{-label}}$ of sentence $\phi$ with respect to database $\Delta$ is the set of all conjunctive clauses $\alpha$ such that $\alpha$ belongs to language $\mathcal{A}$ and $\neg\alpha$ is a minimal supports for $\phi$ with respect to $\Delta$.*

The relation between the $\mathcal{A}$-label of a sentence and its argument is a corollary of the following theorem.

**Theorem 10** *Let $\Delta$ be an argument database with respect to $(\mathcal{L}, \mathcal{A})$. The sentence $\alpha$ is a prime implicant for $\Delta(\mathcal{A})$ precisely when $\alpha$ belongs to language $\mathcal{A}$ and $\neg\alpha$ is a minimal support for sentence $\phi$ with respect to database $\Delta$.*

As the following corollary shows, the $\mathcal{A}$-label of a sentence is simply its argument put in a prime implicant form.

**Corollary 2** *Let $\Delta$ be an argument database with respect to $(\mathcal{L}, \mathcal{A})$. The $\mathcal{A}$-label of sentence $\phi$ with respect to database $\Delta$ is the set of prime implicants for argument $\Delta(\mathcal{A})$.*

**Example 10** Consider the argument database represented by the argument network in Figure 3:

$$OK(X) \wedge A \qquad \supset \quad D$$
$$OK(X) \wedge \neg A \qquad \supset \quad \neg D$$
$$OK(Y) \wedge B \wedge C \qquad \supset \quad E$$
$$OK(Y) \wedge (\neg B \vee \neg C) \quad \supset \quad \neg E$$
$$OK(Z) \wedge (D \vee E) \qquad \supset \quad F$$
$$OK(Z) \wedge (\neg D \wedge \neg E) \quad \supset \quad \neg F$$

The argument for the sentence $\neg A \wedge B \wedge C \supset F$ is $(OK(X) \vee OK(Y)) \wedge OK(Z)$. The prime implicants of this argument are $OK(X) \wedge OK(Z)$ and $OK(Y) \wedge OK(Z)$, each of which is an argument for $\neg A \wedge B \wedge C \supset F$. Moreover, by Corollary 2, these prime implicants constitute the label for the sentence $\neg A \wedge B \wedge C \supset F$.

## 6.3    Diagnosis

The basic task of a kernel–diagnosis system is to compute the kernel diagnoses of an observation with respect to some database. Roughly speaking, a kernel diagnosis of an observation is a "strongest" possible consequence of the observation. More formally, we have the following definition [de Kleer *et al.*, 1992]:

**Definition 13** *The $\underline{\mathcal{A}\text{-kernel diagnoses}}$ of sentence $\phi$ with respect to database $\Delta$ are the prime implicants for the conjunction of all the prime implicates (that belong to language $\mathcal{A}$) of database $\Delta \cup \{\phi\}$.*

The relation between kernel diagnoses and arguments is a corollary of the following theorem.

**Theorem 11** *The conjunction of all the prime implicates of database $\Delta$ that belong to language $\mathcal{A}$ is equivalent to the strongest sentence that belongs to language $\mathcal{A}$ and is entailed by database $\Delta$.*

**Corollary 3** *Let $\Delta$ be an argument database with respect to $(\mathcal{L}, \mathcal{A})$. The $\mathcal{A}$-kernel diagnoses of sentence $\phi$ with respect to database $\Delta$ are the prime implicants for the negated argument $\neg\Delta(\neg\phi)$.*

**Example 11** Consider the database in Example 10. And suppose we observe $\neg A \wedge B \wedge C \wedge \neg F$, which is unexpected given that all gates are OK. Suppose further that we want to compute the kernel diagnoses of this observation. According to Corollary 3, we must first compute the argument for the negated observation. The negated observation in this case is $\neg A \wedge B \wedge C \supset F$, and its argument was computed in Example 10: $(OK(X) \vee OK(Y)) \wedge OK(Z)$. Negating this argument, we get $(\neg OK(X) \wedge \neg OK(Y)) \vee \neg OK(Z)$. The prime implicants of this sentence are $\neg OK(X) \wedge \neg OK(Y)$ and $\neg OK(Z)$. That is, either gates $X$ and $Y$ are not OK, or that gate $Z$ is not OK. Each of these is a kernel diagnoses of the observation $\neg A \wedge B \wedge C \wedge \neg F$.

## CONCLUSION

In this paper, I have identified a logical notion of independence that resembles probabilistic independence. I have also presented independence–based tools to represent and reason with a class of propositional databases that has several applications. The suggested tools have successful counterparts in the probabilistic literature.

## ACKNOWLEDGEMENT

This work was supported in part by grants from the Air Force Office in Scientific Research, AFOSR 90 0136, and the National Science Foundation, IRI-9200918.